\RequirePackage{fix-cm}
\documentclass[smallextended]{svjour3}       
\smartqed  
\usepackage{graphicx}
\usepackage{cite}
\usepackage{multirow}
\usepackage{xcolor}
\usepackage{hyperref}

\usepackage{amsmath}
\usepackage{algorithmic}

\usepackage[caption=false,font=normalsize,labelfont=sf,textfont=sf]{subfig}

\usepackage{url}

\begin{document}

\title{Relational-Grid-World: A Novel Relational Reasoning Environment and An Agent Model for Relational Information Extraction
}

\titlerunning{Relational-Grid-World}        

\author{Faruk Kucuksubasi         \and
        Elif Surer 
}


\institute{F. Kucuksubasi \at
             Department of Modeling and Simulation, Graduate School of Informatics, Middle East Technical University, Ankara, Turkey \\
              Tel.: +90-312-2107888\\
              Fax: +90-312-2102745\\
             \email{faruk.kucuksubasi@metu.edu.tr} 
           \and
           E. Surer \at
             Department of Modeling and Simulation, Graduate School of Informatics, Middle East Technical University, Ankara, Turkey \\
              Tel.: +90-312-2107888\\
              Fax: +90-312-2102745\\
              \email{elifs@metu.edu.tr} 
}

\date{Received: 12.07.2020 }

\maketitle

\begin{abstract}
Reinforcement learning (RL) agents are often designed specifically for a particular problem and they generally have uninterpretable working processes. Statistical methods-based agent algorithms can be improved in terms of generalizability and interpretability using symbolic Artificial Intelligence (AI) tools such as logic programming. In this study, we present a model-free RL architecture that is supported with explicit relational representations of the environmental objects. For the first time, we use the PrediNet network architecture in a dynamic decision-making problem rather than image-based tasks, and Multi-Head Dot-Product Attention Network (MHDPA) as a baseline for performance comparisons. We tested two networks in two environments ---i.e., the baseline Box-World environment and our novel environment, Relational-Grid-World (RGW). With the procedurally generated RGW environment, which is complex in terms of visual perceptions and combinatorial selections, it is easy to measure the relational representation performance of the RL agents. The experiments were carried out using different configurations of the environment so that the presented module and the environment were compared with the baselines. We reached similar policy optimization performance results with the PrediNet architecture and MHDPA; additionally, we achieved to extract the propositional representation explicitly ---which makes the agent's statistical policy logic more interpretable and tractable. This flexibility in the agent's policy provides convenience for designing non-task-specific agent architectures. The main contributions of this study are two-fold ---an RL agent that can explicitly perform relational reasoning, and a new environment that measures the relational reasoning capabilities of RL agents.
\keywords{reinforcement learning \and relational reinforcement learning \and relational reasoning \and relation networks \and attention networks}
\end{abstract}

\section{Introduction}
\label{intro}
Games provide convenient testbeds and experimental environments to model complex scenarios that require sophisticated cognitive abilities. In these environments, unlike everyday life, actions of the decision-making entity called the agent can be measurably rewarded or punished with a reward signal. Using this reward mechanism, the agent can learn to act optimally in the environment. Although this behaviorist perspective does not explain human nature entirely, it is an inspiration for optimal policy search in the field of Reinforcement Learning (RL). In the RL field, agents in an environment are rewarded or punished in terms of selected actions and/or states. This reward mechanism acts as a cost function for policy search, and the agent(s) tries to maximize the cumulative reward. During this optimization process, the agent can search for the state-action mapping (policy) or search for the weighted future reward return of the states (value function). Ideally, the agent(s) will be able to take the best actions in the environment. However, large state spaces and delayed feedback from the environment complicate this optimization problem. During the policy search, the agent should balance their actions based on previously experienced solution paths (exploitation) and not yet experienced paths (exploration). The dilemma here is that in case of excessive exploitation, the agent will never experience the global solution and on the other hand, even if it explores and finds the global solution, it can move to arbitrary solutions. Besides, it is difficult to derive a generalizable policy for different configurations of the same environment. There are various RL methods which have been proposed according to the characteristics of the environments and the performance expected from the agent. The agent, which knows nothing about the environment model, can try to learn the model itself (model-based) by taking actions. However, this method requires the environment to be modeled very well, and the process will be computationally complex as the number of states in the environment increases. For this reason, it is preferable to try to learn the policy and/or value function without learning the environment model (model-free) [1].

Most real-life tasks contain a large number of states. According to classical methods (e.g., Q-learning [2]), huge and difficult-to-create lookup tables are required in order to overcome computational complexity caused by large numbers of states. It has been proposed with [3] that it may be useful to use function approximators instead of lookup tables. Therefore, recent deep learning (DL) methods, which are powerful tools for function approximation, are commonly used. Thanks to large datasets, hardware power and sophisticated DL methods, recent RL algorithms can perform superhuman performance in specific tasks. However, the interpretability, generalization capabilities and data efficiencies of these methods are quite low [4]. These algorithms usually recognize the associations rather than looking for causality in the data. According to Judea Pearl, these algorithms are limited by the capabilities of the curve-fitting [5]. These shortcomings can be overcome using symbolic representation as in classical artificial intelligence (AI) algorithms, given that they can tackle these three problems, but they are not robust and have to be hand-crafted. Therefore, it is clear that a bridge must be established between the symbolic representation and modern algorithms. As Pearl described, the information extracted by the algorithms should be extended to the levels from the “associational information” to the “intervention and counter-factual information.” For this purpose, creating a representation based on the relational information between objects [6, 7, 8] in the environment, as used in symbolic AI, can partially overcome these problems. However, these solutions often do not provide an explicit relational clue. Using PrediNet architecture [9], the relational information between the objects in the environment can be represented explicitly in the post-process, but it seems difficult to use this output in the agent's closed-loop algorithm.

In this study, a new environment called Relational-Grid-World (RGW) \footnote{The source code will soon be released on GitHub.} is introduced. RGW is a two-dimensional (2D) environment where the agent must establish a relationship between the objects in the environment in order to reach the terminal state by getting the optimum reward from the environment. This environment is designed to evaluate the agent's performance in processing relational information. Then, the MHDPA [6] and PrediNet architectures were evaluated in the Relational-Grid-World and baseline Box-World [6] environments. These architectures and environments will be explained in detail in the following sections. The main aim of this study is to assist in the inclusion of causality principles and symbolic mathematics in RL literature. As a result of this study, relational information in the environments was obtained explicitly by using PrediNet, and the agent's policy optimization performance was determined close to the results in the literature (i.e., relation network). This represents a promising outcome given that RL agents need to perform relational reasoning to increase their generalizability and interpretability. In future studies, this study can be elaborated by applying different post-processing methods on explicitly obtained relational information.

\section{Related Research}
The Reinforcement Learning (RL) method offers a mathematical framework to achieve the optimal policy in an environment where the agent interacts [1] with the environment via actions, and gets rewards for the state transitions due to actions. RL is mostly used in sequential decision-making problems, and the agents try to maximize the expected cumulative reward. In model-free problems, Monte-Carlo Tree-Search (MCTS) [10] and Temporal Difference (TD) [1] are the most used methods due to their ability to work independently from the environment model. In MCTS, observations have high variance and low bias, while TD methods have the opposite; low variance and high bias characteristics. Recently, TD method has become more widespread than MCTS since it is combined with deep neural networks due to its low variance property. During the agent-environment interaction, if the policies that agents behave and estimate are different, this is called off-policy learning. Q-learning [2] can be given as the basic example of off-policy algorithms. The biggest problem with this method is that it can be unstable, but there are some tricks such as Experience Replay [3] to prevent instability. On the other hand, when the behavior policy and learning policy are the same, it is classified as an on-policy algorithm. SARSA [1] can be given as the basic on-policy example. The chronic problem of on-policy methods is that they may not reach the optimum policy. Apart from the policy classification of algorithms, RL algorithms can be divided into two as value-based and policy-based. In value-based methods, the values (expected cumulative reward) of the states are tried to be estimated as in Deep Q-network (DQN) [3]. Prioritized Experience Replay [11] method has been developed for the DQN method to experience replay more efficiently. In policy-based methods [12], an optimum policy is tried to be obtained directly. The biggest advantage is that they can be used in continuous action spaces. In addition, they can converge faster than value-based methods, but they are less likely to reach global optimality. There are also actor-critic [13] algorithms that try to merge the advantages of these two methods. These algorithms perform an approximation for value function and try to optimize their policies using this approximation. Actor-Critic methods can be distributed to multiple agents in order to collect high variance samples and speed up the learning process [14]. Thanks to the Importance-Weighted Actor Learner Architecture (IMPALA) [15] algorithm, parallel learning can be done more efficiently with a fast and scalable policy gradient agent and V-trace correction method.

These powerful model-free algorithms/frameworks can overcome the decision problems that a person can solve quickly by using too many samples. For this reason, it will be more reasonable and also challenging to search the policy through high-level features by making temporal abstraction in the environment. When appropriate temporal abstraction can be made, the solution can be reached by using systematic planning and control in the environment using Hierarchical RL Methods [16]. Apart from that, delayed sparse reward signals in the environment is also a big problem for deep reinforcement learning problems. For example, Montezuma Revenge is an environment that reflects this problem. The algorithms that solve the Montezuma are quite environment-specific methods [17]. It has been observed that their performance can be increased by creating intrinsic curiosity in such environments [18]. Moreover, creating generalizable, interpretable, and transferable knowledge by the agent is a much more theoretical problem in the RL domain.

The Self-Attention mechanism, which is also the method used in this study, is widely used for sequence-to-sequence modeling in natural language processing (NLP) problems [19], and studies show that they can be boosted with multi-head operations [6]. This mechanism can also be used in the RL domain due to its ability to extract relational information from the data. The use of relational information in RL problems has been proposed in the past [20], which can be applied more effectively with current deep neural network methods. Relation Networks (RN) have been established by combining attention mechanisms with current up-to-date Deep RL methods. By using the relation-based methods, the relevance between the units (or objects) in the sensory input can be extracted, and this provides a more efficient learning representation. Relation networks seem to be a promising method in terms of interpretability and generalization. Firstly, in the [7, 8] article, these modules were used to extract the relation between the objects from the images. Later, RN was used with [21] to increase the performance for language modeling. Then, [21] was used for boosting the decision-making capability of the agent in a dynamic environment. These networks provide easy-to-interpret visual clues. However, it is important to be able to clearly reveal the relationship between the objects in order to use symbolic mathematics. For this purpose, in PrediNet [9], explicit information about objects is derived from images by using RN, and explicit information can be post-processed with logic programming languages. In this study, we test the same method in two different environments —i.e., our proposed Relational-Grid-World environment and Box-World as the baseline.

\section{Methodology (Architecture)}
In this study, for the sake of computational efficiency, agents were trained with Asynchronous Advantage Actor-Critic (A3C) framework [14], which is a parallel actor-learners method in the Deep RL domain. The agent(s) derives a latent space by taking RGB input from the environment and predicts the actor function (policy logits) and critic (baseline value function) from the latent space by using two different Multilayer Perceptrons (MLPs). This estimation is performed in parallel by multiple asynchronous agents, and a global network is trained. The baseline value function model is trained using the temporal difference method, and it is used as a reference for training the estimated policy logits. In addition to the loss function used for optimizing the value and policy functions, the entropy of the policy is also added (Eqn. 1) as in [14]. In this way, the balance between exploration and exploitation can be adjusted more precisely. Gradient updates of the global network are made when an episode ends, or the n-step buffer is full.
\\
\\
\\
\\
\noindent Actor loss function:

$d \theta \leftarrow d \theta+\nabla_{\left(\theta^{\prime}\right)} \log \pi\left(a_{i} | s_{i} ; \theta^{\prime}\right)\left(R-V\left(s_{i} ; \theta_{v}^{\prime}\right)\right)$ \\

\noindent Critic loss function:

$d\theta_{v} \leftarrow d \theta_{v}+\partial\left(R-V\left(s_{i} ; \theta_{v}^{\prime}\right)\right)^{2} / \partial \theta_{v}^{\prime}$ \\

The raw sensory RGB input from the environment was passed through two convolutional layers with 24 and 12 kernels. The positional information of each pixel (x,y) is added to the Convolutional Neural Network (CNN) output as an additional channel and is sent to the core module. In the agent core module, relational modules and PrediNet modules are tested separately. The full pipeline of the agent's network architecture can be seen in Figure 1. Input and output array sizes are identical for both  Multi-Head Dot-Product Attention (MHDPA) and PrediNet models. Moreover, there is a switch block after the relational module to exchange network models for different experiments. Hyperparameter sets of both architectures can be found in Appendix Table 2 and Table 3.

\begin{figure}
\centering
\includegraphics[width=2.5in]{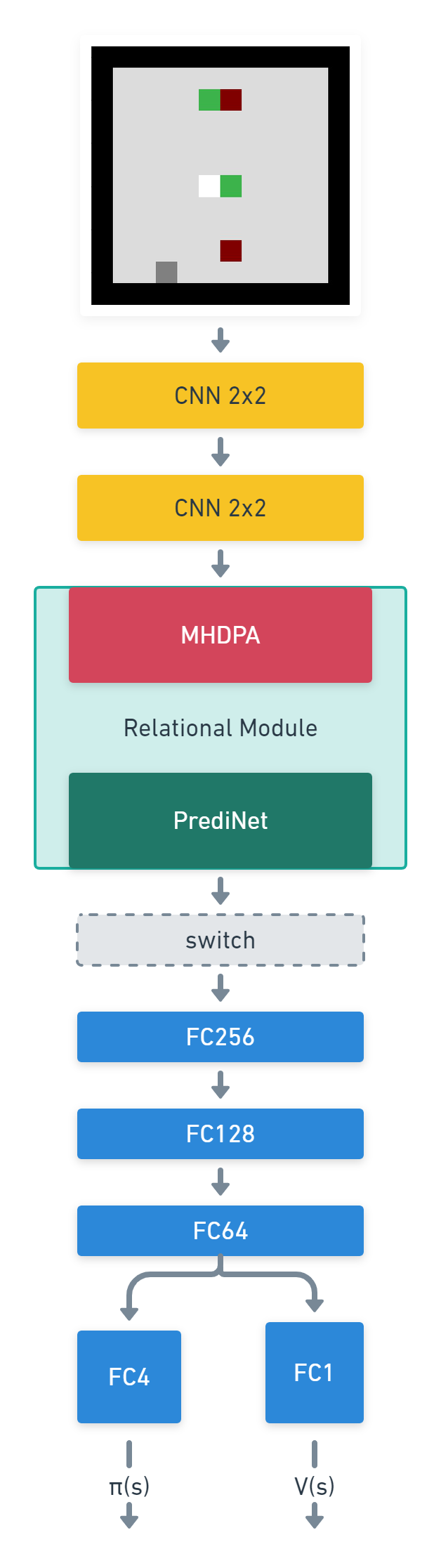}
\DeclareGraphicsExtensions.
\caption{Agent network architecture, switching between PrediNet and Multi-Head Dot-Product Attention modules.}
\label{fig1}
\end{figure}

\subsection{ Multi-Head Dot-Product Attention (MHDPA) Module}
The MHDPA module is applied as used in [6]. All data coming as CNN output are flattened in the direction of positional dimension (\emph{E}), and the linear transformation is done with query ($\boldsymbol{W}^{q}$), key (${W}^{k}$) and value ($W^{v}$) trainable weights. The transformed matrices (relatively named as \emph{Q} (query), \emph{K} (key) and \emph{V} (value)) are compared with dot-product and scaled with the dimension of the key attention matrix ($d_{k}$). Then, softmax operation is applied to the output and weighted with \emph{V} matrix. \\

\noindent Attention Formula:

$A_{H}(\boldsymbol{E})=\operatorname{softmax}\left(\frac{E \boldsymbol{W}^{q}\left(E W^{k}\right)^{T}}{\sqrt{d_{k}}}\right) E W^{v} \quad, \\ \text { where } \: \boldsymbol{ H}: \text { head index }$ \\ 

In this way, the weight information of all entities on each other will be stored in the \emph{Q} and \emph{K} matrices. In this process, the data is broken up by the number of heads and the same calculations are made in parallel for each piece with different weight sets. Finally, the output (the \emph{A} matrix) is passed through a feature-wise max pooling and two-layer multilayer perceptron, and policy logits and baseline value functions are estimated.

\subsection{PrediNet Module}
Unlike MHDPA, in the PrediNet module [9], \emph{Q} matrices are estimated differently for each head. Therefore, the network can calculate the same relation set of two different units in each head. In this way, a global relation function is obtained, and the relations of the units can be transformed into the same base representation. In the experiments, 32 different relation bases were used. The main difference of this method from the MHDPA is that, by using the estimated global relation function, the relation values for each pair object become comparable to each other. The final output of the network with \emph{k} heads and \emph{j} relation is that: \\ 

\noindent Object Relation Function:

$\psi_{\mathrm{i}}\left(d_{\mathrm{i}}^{h}, e_{1}^{h}, e_{2}^{h}\right), \quad \text { where\: } h<k, \quad i<j$\\ 

$d_\mathrm{i}^{h}$ term is the abstract relational distance between $e_{1}^{h}$ (entity/object 1) and $e_{2}^{h}$ (entity/object 2) in an arbitrary head. The distance information can be used in downstream processes. Since the original PrediNet was not used in the original RL problem, in order to estimate the policy logits and baseline value function, two additional linear transformations are made at the end of the architecture.

\section{Environments}

\subsection{Baseline Environment: Box-World}
Box-World (Figure 2), defined in the [6], contains single-colored boxes in pairs in a 12x12 pixel environment. The agent can go up, down, right and left directions, and it can collect a box by standing on it. The colors of each adjacent block pair are different, and there can be another box that is identical to one of the pairs. The color of the right box represents “locks,” and the left box color represents the “keys” for a pair. The “agent” box (grey colored) can retrieve another box by standing on it only if it is free. Also, key boxes can only open lock boxes of the same color, and the opened lock boxes release the adjacent key box. At the beginning, a free key is generated in order to avoid deadlock situations. The ultimate goal is to access the “gem,” which is a white colored box. When the agent reaches the gem box, the game is terminated and the task is completed.

In each episode, there is a unique solution, and there are distractor branches leading to dead-ends. In this environment, the agent enters a randomly generated environment in each episode. Agents should notice whether a box is on a distractor or a solution path. Also, it is necessary to solve the relationship between the boxes in the environment because key-lock couples are randomly located in the environment. The level of difficulty of the environment can be adjusted by increasing the solution length, the number of distractor branches, and the length of the distractor branches in the environment. The probability of finding the correct solution by chance is very low (2.3\%).

Unlike the test environments generated in Relation Networks, visual information of the environment is reduced to 12x12 pixels instead of 14x14. This reduction makes it possible to use limited hardware resources more efficiently. Unlike the baseline usage, the agent is considered to have received the new key/gem block without having to visit the key block as soon as the agent opens the adjacent lock box. This difference has no effect on the overall conclusion since it does not affect the relational information between the objects, and this statement is valid for both algorithms tested.
\begin{figure}
\centering
\subfloat[Configuration-1: one key/lock pair (solution length is 1).]{%
\includegraphics[clip,width=2.5in,height=2.5in]{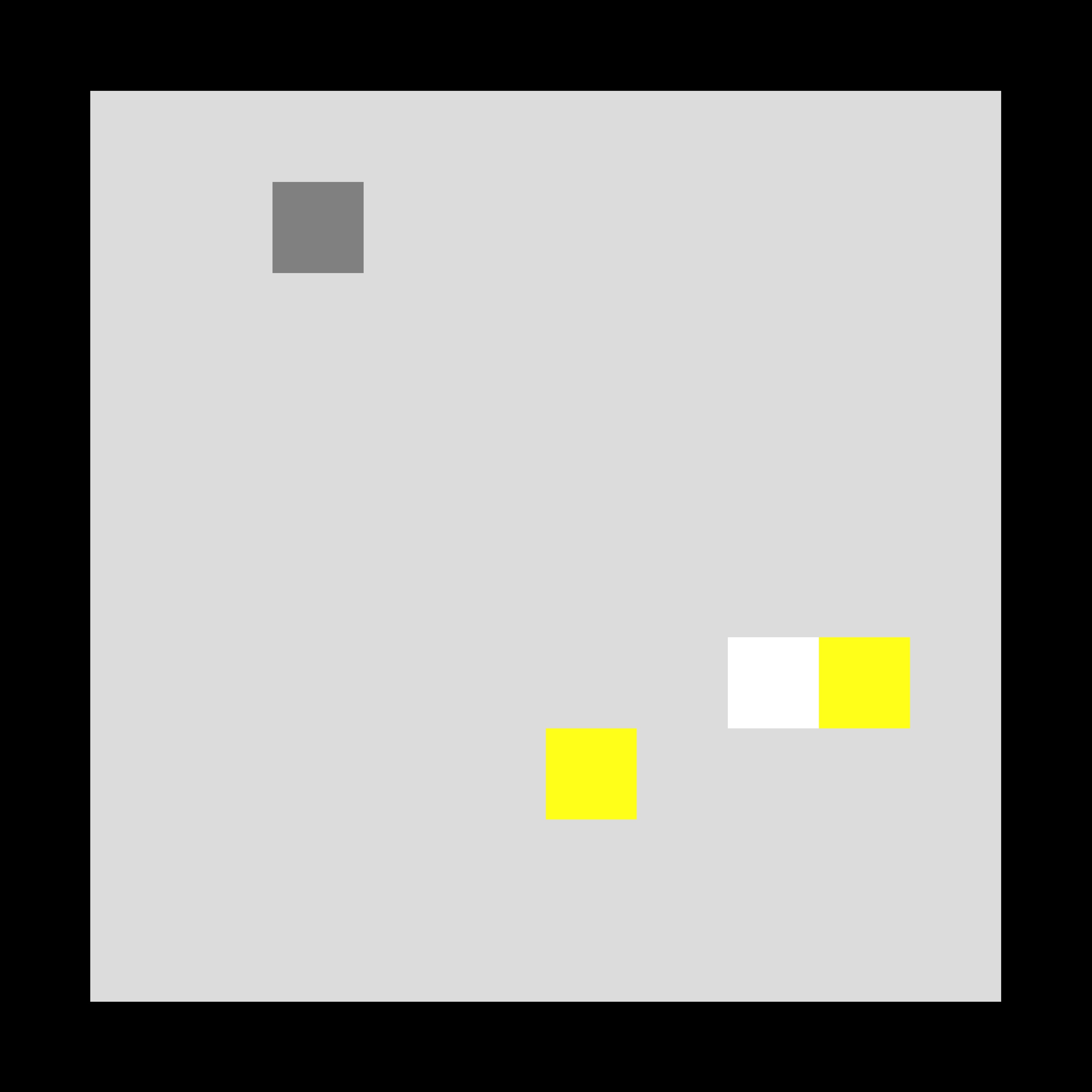}%
}

\subfloat[Configuration-2: one key and two locks (solution length is 1 with a distractor block).]{%
\includegraphics[clip,width=2.5in,height=2.5in]{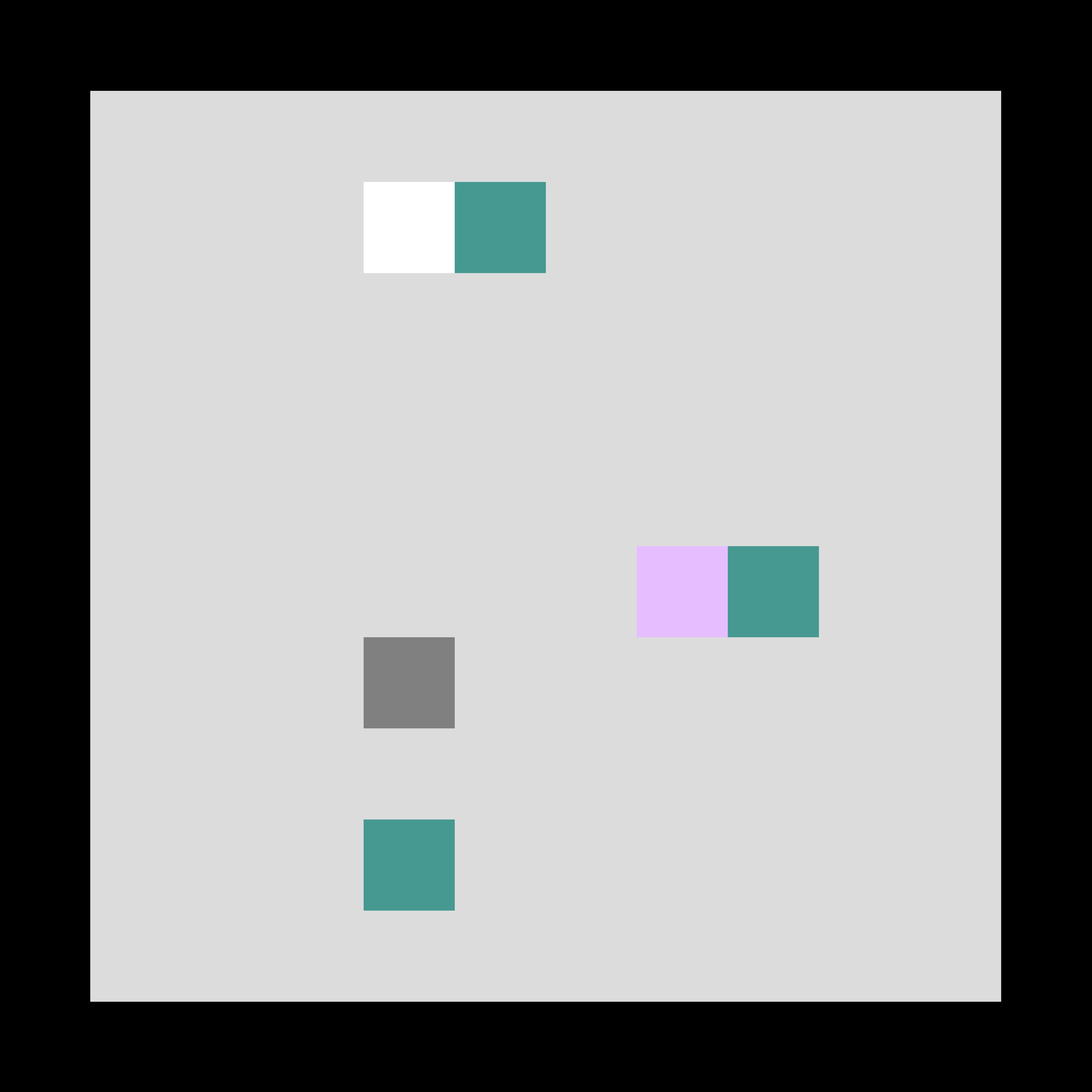}%
}

\subfloat[Configuration-3: two key/lock pairs (solution length is 2).]{%
\includegraphics[clip,width=2.5in,height=2.5in]{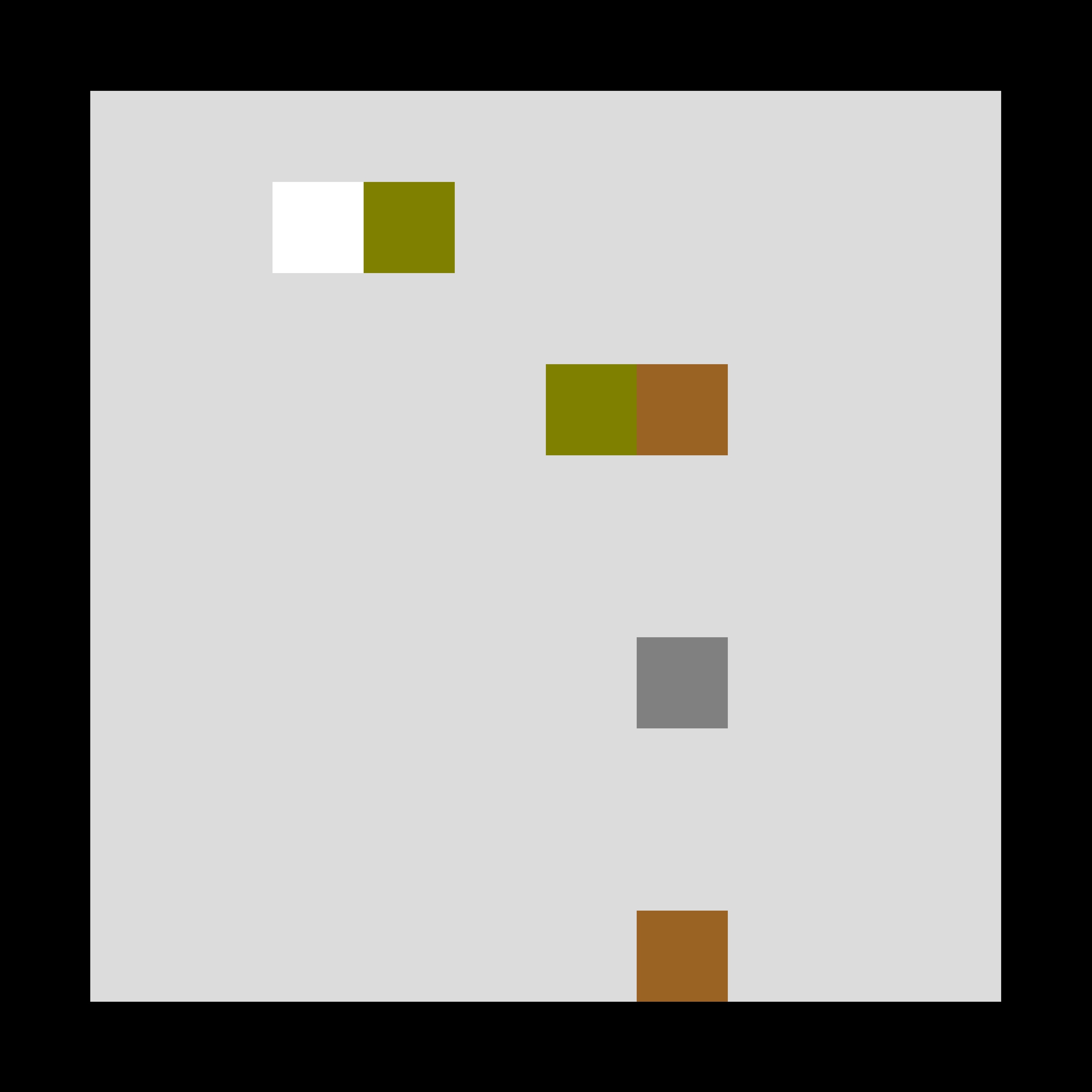}%
}

\caption{Randomly generated Box-World Environments (Agent: Grey, Gem: White, Key/Lock: Other Colors).}

\end{figure}

The configurations have been arranged to reflect the shortest basic problems in the environment. This is because core modules are tested under hardware limitations. In the first configuration experiments, there is only one key-lock pair. Hence, it is possible to compare the learning speeds of the two methods in the simplest case and to see what the upper limit of the PrediNet algorithm is. In the second configuration, additionally, there is a distractor branch that leads to a dead-end. Therefore, the algorithms will need to learn which blocks should be avoided or not. In order to succeed on this task, the agent will need to distinguish between the “gem” box and “distractor” box, and their path by backtracking. Finally, in the third configuration, there is no distractor branch, but there are two key-lock pairs. Thus, the third configuration tests the algorithm's ability to establish multiple sequential relational information. In order to limit the training period, environments were terminated after 300 steps of the agent.

\subsection{Relational-Grid-World (RGW)}

In this study, we introduce a new environment, Relational-Grid-World (RGW), which is complex in terms of visual perceptions and combinatorial selections. RGW has 10x10 pixels and contains eight objects, which can be regenerated procedurally (Figure 3-b). It is fully observable, and the agent can go up, down, right and left directions one grid at a time. The complexity of the environment can be adjusted by playing with the state space size (grid size) and the number of repetitions of the objects. The interdependent objects must be used in the correct order by the agent when necessary in order to solve the environment in an optimum way. The reward functions used for the two environments can be seen in Table 1. While determining the reward values of the Relational-Grid-World environment, an analogy was established with the Box-World environment. In this way, it is ensured that the network parameters are not different for the two environments.

There are two terminator objects; the pit \footnote{Pit icon, Non-commercial use: \url{https://favpng.com/png_view/circle-spiral-circle-clip-art-png/wLy3E82g}}, and the terminal \footnote{Terminal icon, Non-commercial use: \url{https://www.flaticon.com/free-icon/plug_31863}} (Figure 3-a). The reward of the terminal state is defined relatively high to other objects (+10) to ensure that the correct solution is unique in terms of visited objects sequence. For the correct solution of the task, the agent is always expected to reach the terminal state by finding the optimum path in the current configuration of the environment. Terminator objects can be seen as equivalent to gem objects in the Box-World environment. The location of these objects in the environment configuration is the primary factor affecting the difficulty of the solution. Therefore, it will be appropriate to position the objects after determining the location of the terminal object during the creation of the task. The pit object is one of the objects which gives the largest penalty (-1) to the agent. In cases where the terminal state cannot be found by the agent or it does not exist, the pit is a secondary solution for preventing the infinite penalty when there is another object giving a negative reward. When the agent's exploration ability is not enough, the agent will try to finish the episode through the pit object instead of the terminate object. Therefore, the pit object is a useful tool for understanding the balance between the agent's exploration and exploitation behavior.

The two most crucial objects in the environment are the enemy \footnote{Villain icon, Non-commercial use: \url{https://www.pngegg.com/en/png-nfhkb}} and sword \footnote{Sword icon, Non-commercial use: \url{https://www.pngwing.com/en/free-png-nxuvd}} objects (Figure 3-a). They are two objects with the strongest connection in the environment because the reward received from the enemy object varies according to the agent-sword object history in an episode. When the agent reaches the enemy state, it is penalized with -1 point, while the only way to escape from this penalty is to reach the sword state in advance. In some generated environments, the enemy is on the optimal path, so it is critical to visit the sword in advance. However, the enemy can be a dummy state, and the agent is expected not to go to the sword state unless necessary. In some episodes, the agent must get the sword object before reaching the enemy object, while in another episode, the agent can solve the task without taking the sword. Therefore, the agent's understanding of the strong relationship between these two objects is an important task for the agent in order to solve the task optimally. Also, there can be other objects on the path from sword to enemy object. These intermediate objects can be seen as distractors (blocks the optimum solution) for the agent's understanding of this relation between these two objects. By changing the number of these distractor objects by the user, the robustness of the agent's relational reasoning ability can be tested. The sword and enemy objects can be used multiple times in an episode. Multiple-use of the sword object eases the solution while increasing the number of the enemy object makes the solution more difficult. Thus, sword and enemy objects are the two most important tools for measuring the relational reasoning capability of the agent in the RGW environment.

\begin{figure}
\centering
\subfloat[RGW environment Objects.]{%
\includegraphics[clip,width=2.5in,height=2.5in]{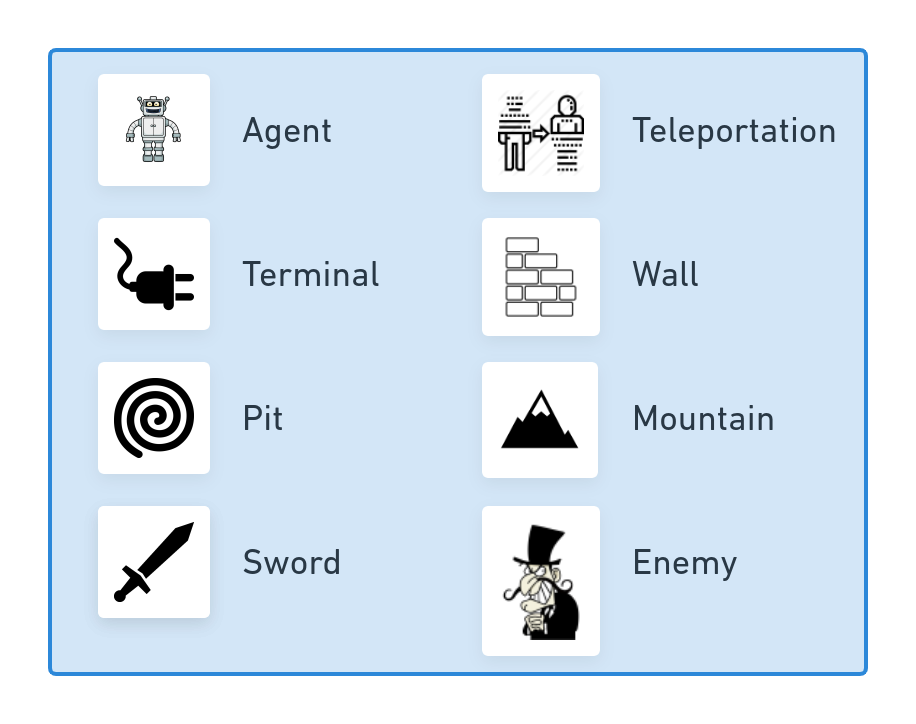}%
}

\subfloat[A simple RGW configuration without any negative rewards.]{%
\includegraphics[clip,width=2.5in,height=2.5in]{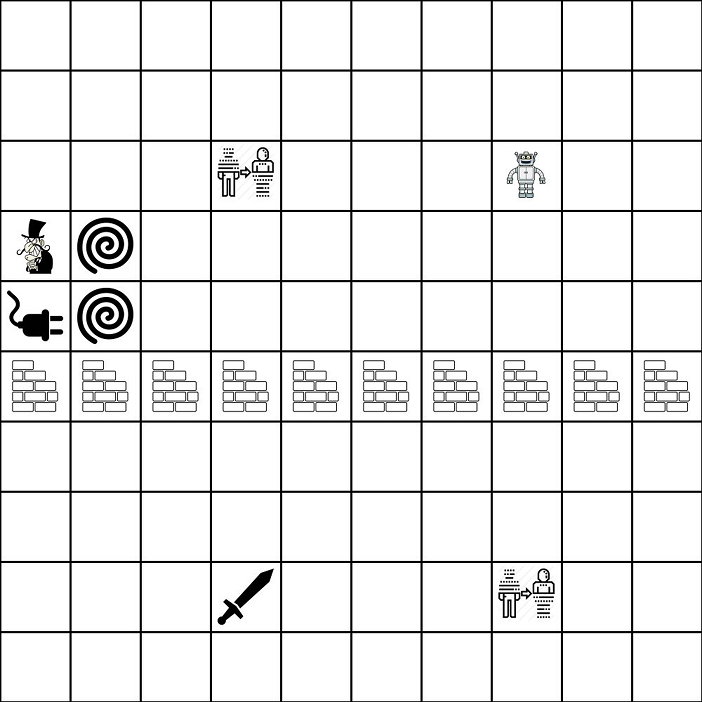}%
}

\subfloat[An RGW configuration which uses all basic features of the objects.]{%
\includegraphics[clip,width=2.5in,height=2.5in]{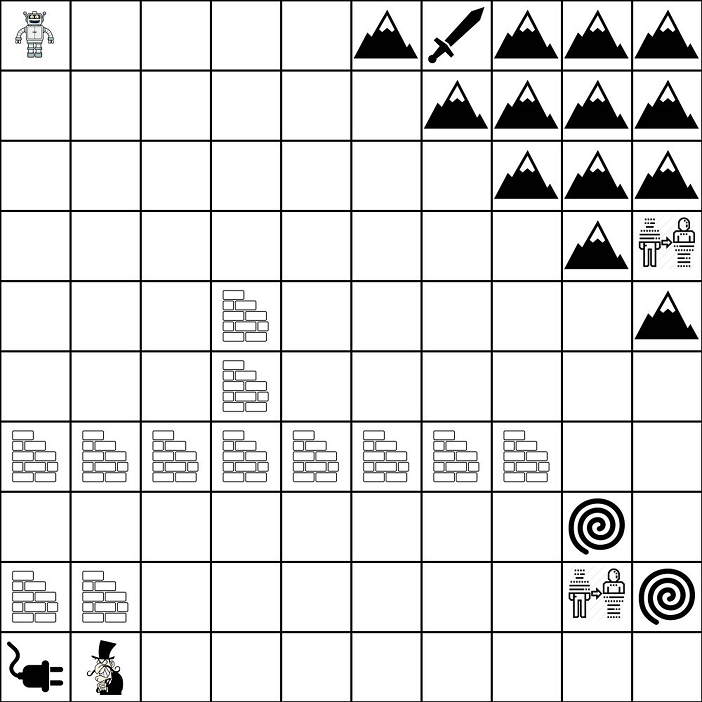}%
}
\caption{Relational-Grid-World (RGW) environment objects and example configurations.} 
\end{figure}

\begin{figure}
\centering
\subfloat[A 20x20 RGW configuration that requires precise position control in a large state-space.]{%
\includegraphics[clip,width=2.5in,height=2.5in]{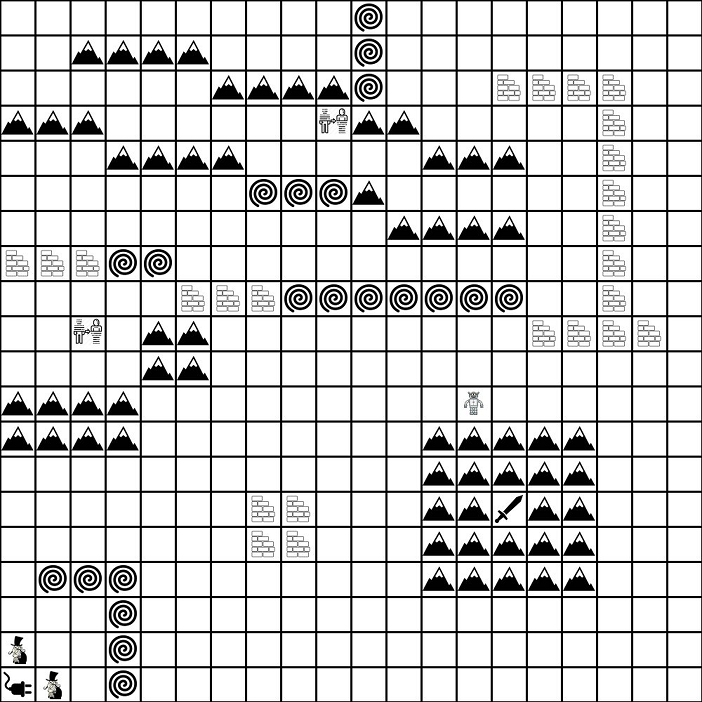}%
}

\subfloat[An RGW configuration with a high negative reward potential.]{%
\includegraphics[clip,width=2.5in,height=2.5in]{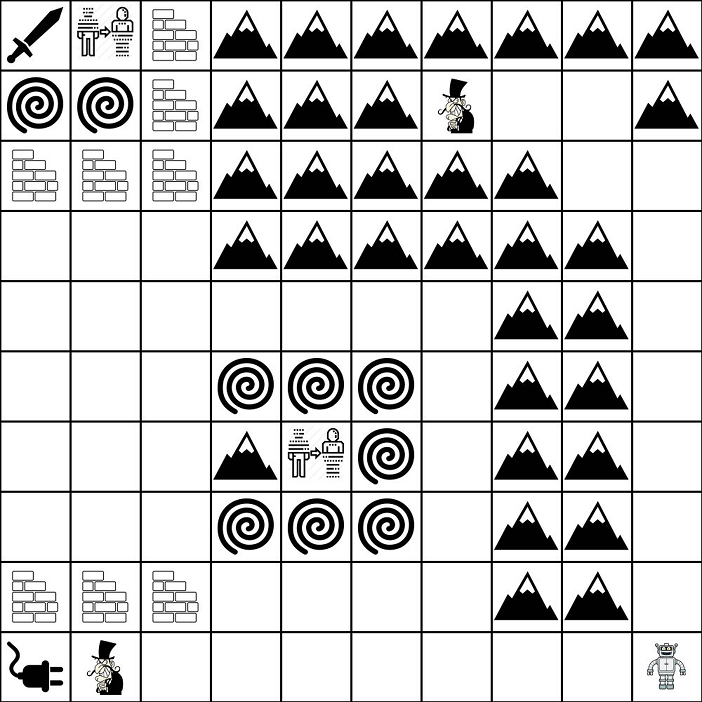}%
}

\subfloat[An RGW configuration that requires relational reasoning on one-to-many objects.]{%
\includegraphics[clip,width=2.5in,height=2.5in]{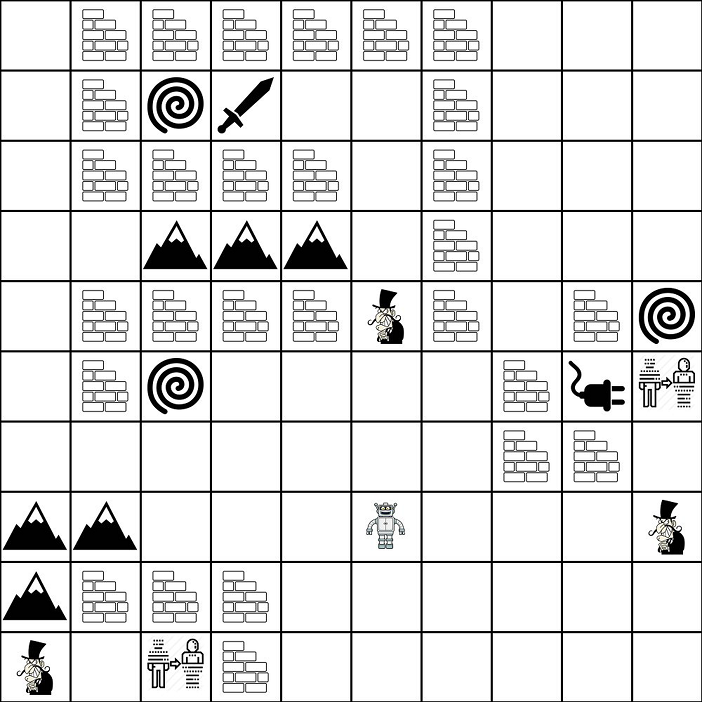}%
}
\caption{Different RGW configuration examples.} 
\end{figure}

Apart from the basic objects, there are three more objects (Figure 3-a) to help shape the solution path as desired. These are the wall \footnote{Wall icon, Non-commercial use: \url{https://favpng.com/png_view/brick-clipart-rectangle-square-brick-clip-art-png/XhH1JjMX}}, mountain \footnote{Mountain icon, Non-commercial use: \url{https://www.hiclipart.com/free-transparent-background-png-clipart-stcbr}}, and teleportation \footnote{Teleportation icon, Non-commercial use: \url{https://pngio.com/images/png-a1788695.html}}, objects. The wall object does not generate any reward signal (0); it only restricts the agent's motion space in the environment. Using this object, it can be made more difficult/easier for the agent to access basic objects, so the wall can indirectly play with the complexity of the task. When the agent visits the mountain object, it gets a reward of -0.01 points. This reward value can be seen as a small penalty (relatively) in the environment. By using the mountain object, the optimum path can be shaped like the wall object. Moreover, it can be placed in an area between the sword and enemy objects, and it acts like a distractor object. In this respect, it is appropriate to determine the sword and enemy positions before determining the position of the mountain object(s). The last object that can be used in the environment is the teleportation object. There must be at least two of the them in the RGW environment (entrance and exit, interchangeably). Thanks to these objects, the agent can go from one grid to another in a single time step. The optimum path can be shaped using these objects, but their main purpose is to measure the agent's sensitivity to the position change. Therefore, the robustness of the agent control algorithm to the dramatic changes of the position information, can be measured.

Two different configurations (Figure 3-c) were used for RGW environment experiments. The only difference between them is that there are no penalty objects (mountain and pit) in the first configuration. In this way, it will be seen how the algorithms will respond to change on the number of penalty objects.

\begin{table}
\renewcommand{\arraystretch}{1.3}
\caption{Reward function analogy between two environments}
\label{table6}
\centering
\begin{tabular}{ccc}
\hline
\multirow{2}{*}{} Box-World & RGW & Reward \\
  Object & Object & Signal\\
\hline
Gem  & Terminal & 10\\
\hline
Key  & Using Sword & 1\\
\hline
Distractor & Enemy/Pit & -1 \\
\hline
- & Mountain & -0.1 \\
\hline
\end{tabular}
\end{table}

\section{Results and Discussion}
The training process took longer than the reference article as clock time due to the use of A3C as the RL framework rather than A2C or IMPALA. However, as stated in the [6], using A3C has no effect on the results, because the only difference is the used parallel training framework, not the agent architecture. The number of the parallel actors used was kept at the maximum value according to the thread number of the CPU hardware.

When PrediNet architecture is trained with a small number of relational representations, it was seen that it could not reach a stable level of performance. Therefore, 32 representations were used in the experiments instead of the eight representations used in the original article. With the increase of the buffer size, it was observed that the training process accelerated. Therefore, the buffer size used during the experiments was selected to reach the upper limit of the GPU memory used. In order to prevent dramatic network updates of the modules, a gradient clip was applied to weight gradients. It was observed that both architectures could not be optimized in cases when the clip value is small. In PrediNet architecture, the relational representation is determined with the subtraction operator (vector difference) by default. When the absolute operation is applied to the vector difference, the model diverged. Similarly, the divergence has occurred when using the sums of squares operator as an alternative to the subtraction. For these experiments, it can be concluded that relational representation values are also dependent on their signs.

\subsection{Box-World Experiments}

Experiments were carried out by using two algorithms in three different environment configurations (Figure 2). The same number of heads (4) were used for both algorithms. In addition, 32 different relation representations are used in PrediNet (PN) architecture. Successfully finished episodes for two modules can be seen in Figure 5. Relation Network (RN or MHDPA) performs better than PN for both solution lengths. There is also a bias between the solution success of modules. Since RN creates a different query and key matrices for all heads, it can use more information about the environment than PN. Therefore, the RN is expected to have better performance. However, despite the lower performance of PN, it gets more critical knowledge with explicit relationship information between the objects in the environment, because it is more suitable for post-processing and it is interpretable. In addition to this, PN trains much faster due to its simpler architecture.

\begin{figure}
\centering
\subfloat[The success of the agent on Box-World configuration 1.]{%
\includegraphics[clip,width=5in,height=2.02in]{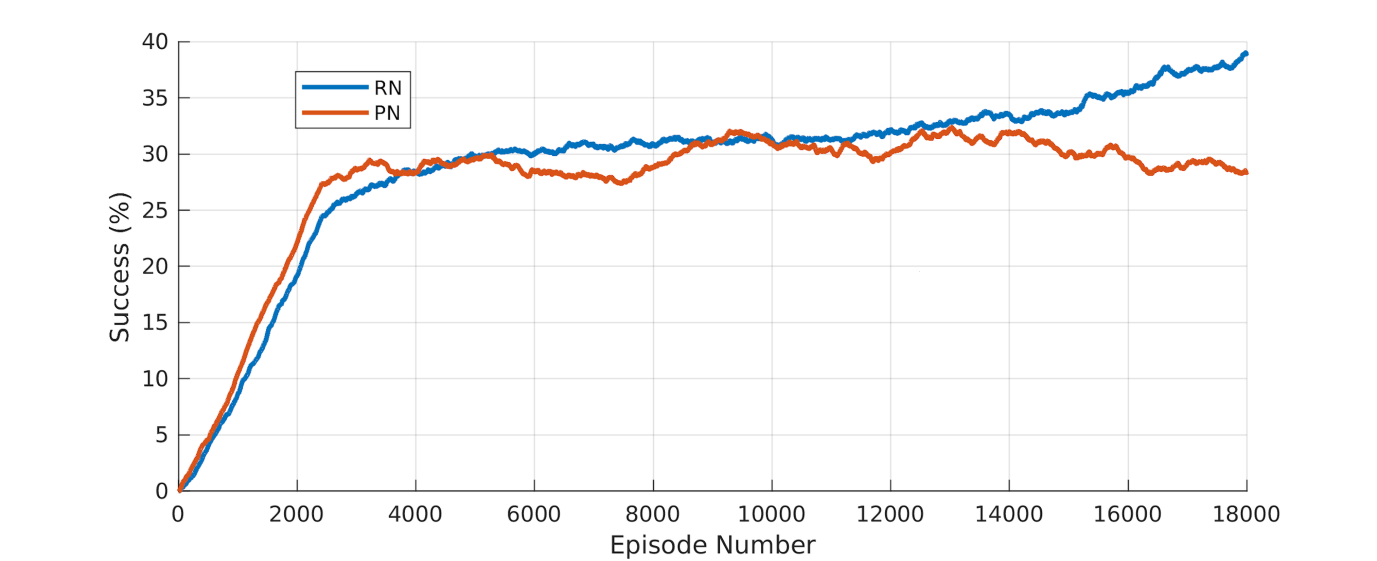}%
}

\subfloat[The success of the agent on Box-World configuration 2.]{%
\includegraphics[clip,width=5in,height=2.02in]{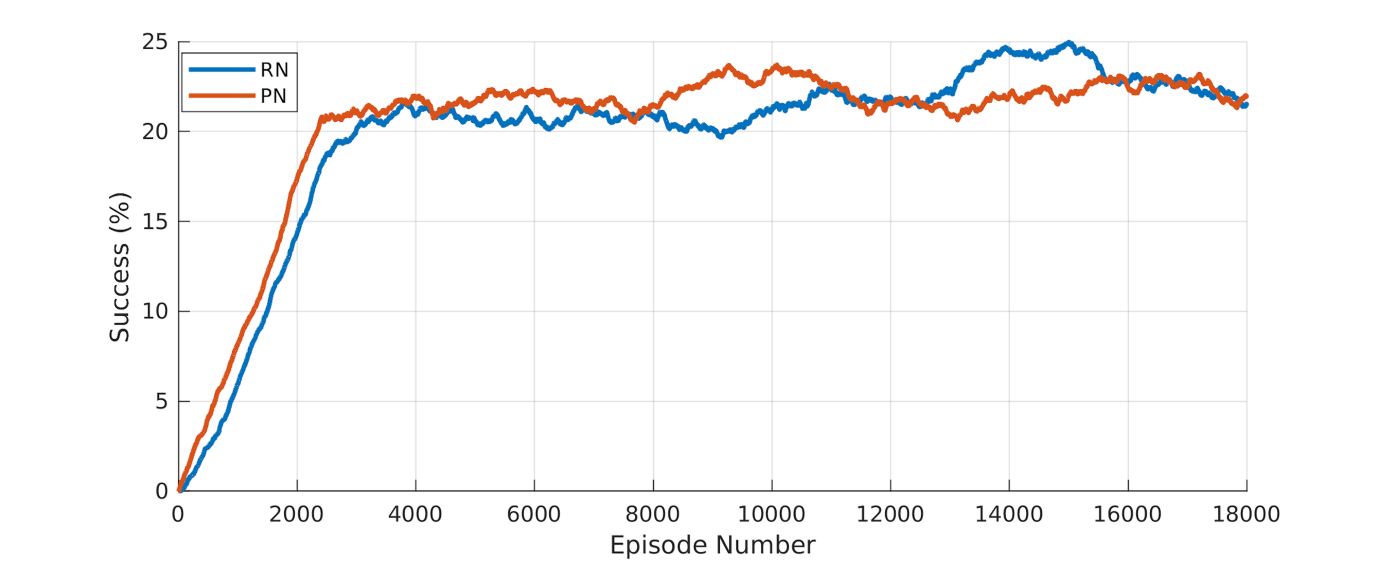}%
}

\subfloat[Percentage success of the agent on Box-World configuration 3.]{%
\includegraphics[clip,width=5in,height=2.02in]{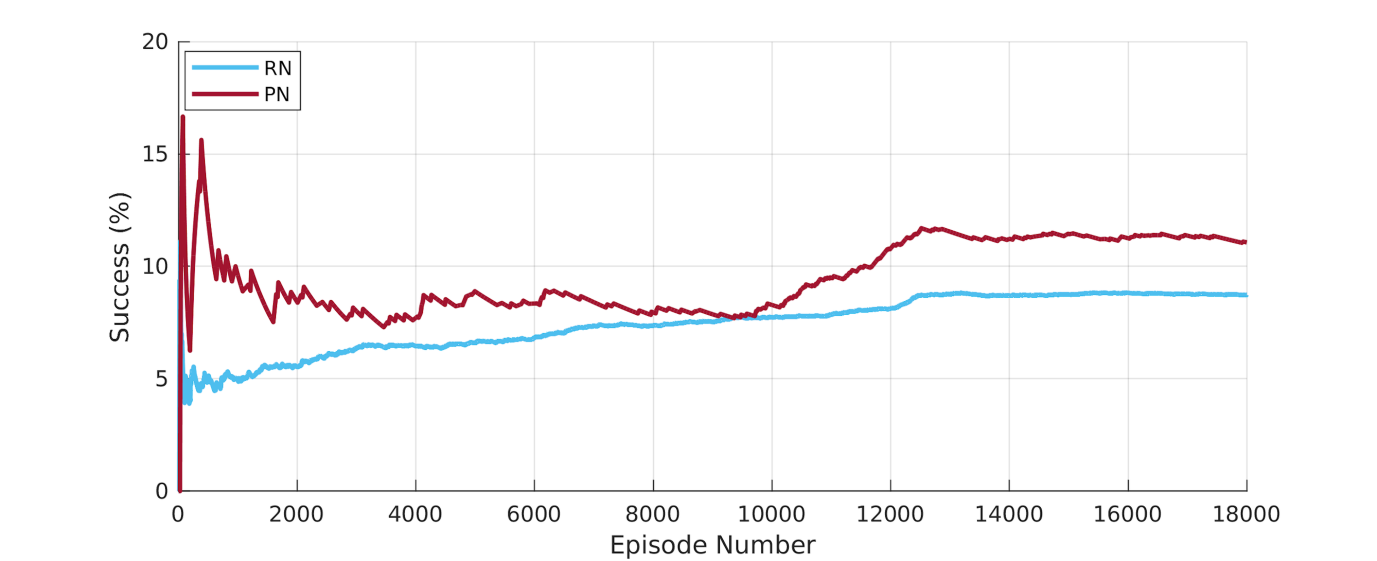}%
}

\caption{Box-World environment performances of PN and MHDPA (RN) modules.} 
\end{figure}

The success percentages in Figure 5 have been determined by the number of the episodes terminated by reaching the gem block in the last 1000 episodes. Considering the number of frames seen by the agent and success rates, it is seen that the PN module performs better than the RN module for a short time, but then the performance of the RN module surpasses with time. Also, for different configurations, there is a big change in the performance of the modules. The most important reason for this is the change in the complexity of the connection between objects. As expected, the percentage of success decreases by one third as the solution length increases. Also, the module's performances are better at one distractor configuration than the two-step solution configuration, given that the agent has to make longer backtracking for the two-step solution. This requires a more complex relational representation of the objects, so the agent performs worse at a two-step configuration than the distracting one. It is known through the outcome of the RN module that achieved full success in these three configurations as a result of longer training times. It is understood that the PN module performs close to RN in the 0 to 18,000 episodes range where it is trained. The performance of the agent trained with PN is based on simpler network complexity and its different object relations information representation. The fact that this representation performs relatively close to RN when the complexity of the environment increases shows robustness with relational complexity.

Figure 6 shows an attention heatmap of a randomly generated configuration-1 Box-World environment. This heat map is extracted from the softmax operation output of the MHDPA network's second head. According to the heatmap, there is strong attention from agents to the free red key. Also, agents have self-attention because the location of the agent is always a critical state in the task. The agent also attends to empty grids, which are around the key and agent objects, because the kernel sizes of the CNN layers are larger than one. As the training continues, the agent's focus will completely shift to the objects, and attention to the empty grids will vanish.

\begin{figure}
\centering
\includegraphics[width=3.5in]{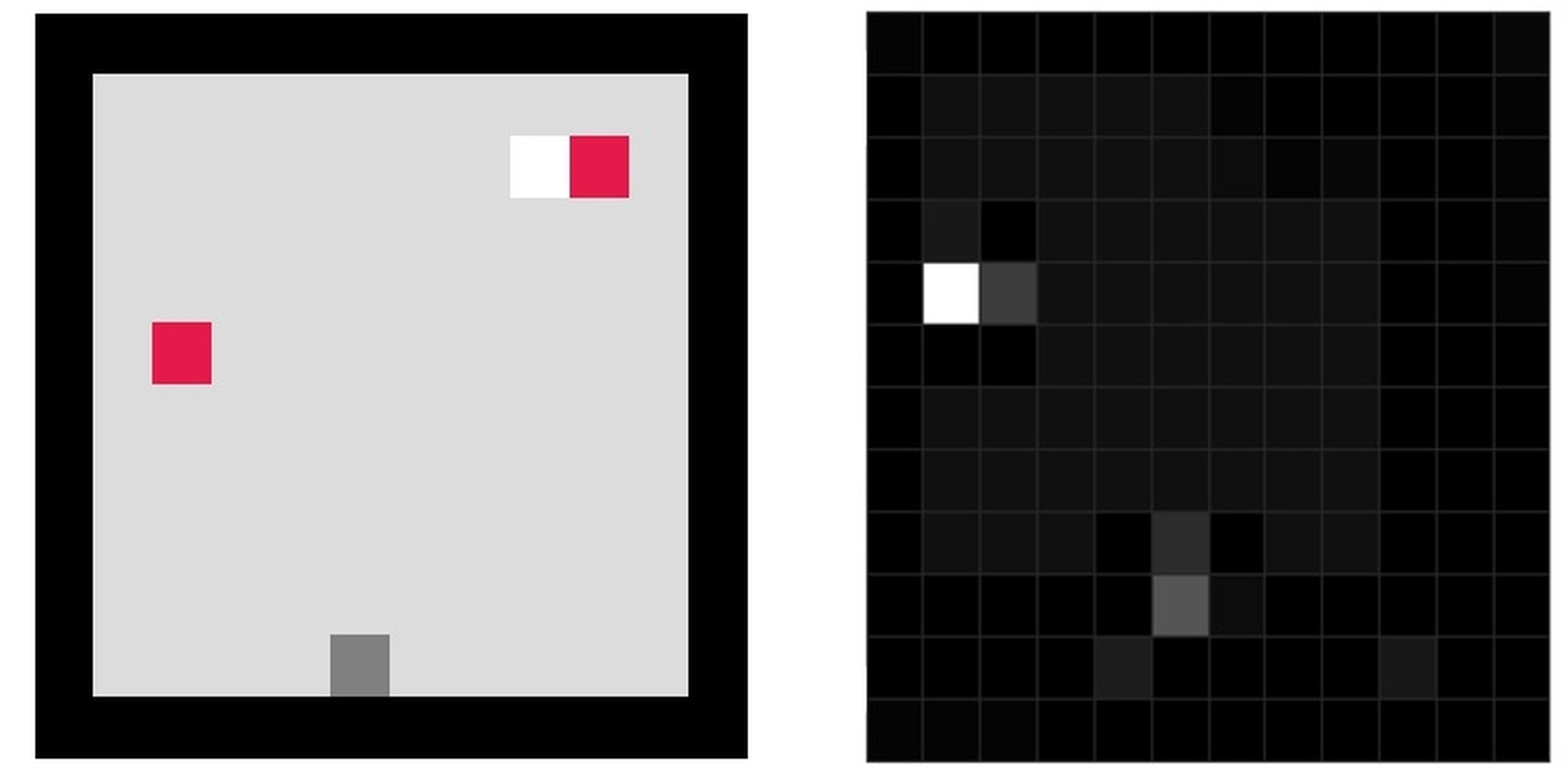}
\DeclareGraphicsExtensions.
\caption{A random configuration-1 Box-World state and related agent attention heat map.}
\label{fig1}
\end{figure}

\subsection{Relational-Grid-World Experiments}

In the default configuration of the RGW environment (Figure 3), two modules are trained with around 60,000 episodes. Training time in this environment has progressed much slower than the Box-World environment. This is because the environment contains more complex visuals and tasks than the Box-World environment. With the set of hyper-parameters (in Appendix, Table 2 and Table 3), both algorithms lose their initial performance over time. One of the reasons for the decrease in the performance is that the network remains at a local minimum. In cases where the reward and penalty points are extreme in the environment, the agent has to go through the states where there are high penalties to reach the terminal state. Therefore, instead of taking the risks of the exploration, the agent starts walking against the walls in the environment and receives neither reward nor punishment. In such cases, the maximum episode length has been determined as 200 steps in order to limit the training time, which did not make any differences in the optimum solution of the environment but helped to the efficient usage of the resources. In addition to this change, the reward function of the environment should be determined correctly. The Box-World environment was used as a reference in the reward function determination process. The reward function was determined by establishing an analogy between the two environments, as in Table 1.

The use of reward analogy enabled close values to be used as hyper-parameter sets in networks. As reward values given in penalty and reward states are close to each other in the reward function, the training process gets quite long. Although this situation can be overcome by exploration/exploitation balance during training, it is quite sensitive to the parameters. The entropy value of the policy logit is added to the loss function of the agent in addition to policy and value loss to ensure the balance of exploration/exploitation. Since the agent will explore as the entropy value rises and exploits as the entropy decreases, its weight in the loss function provides control over this balance. If the exploration effect is kept low, the state of converting to the local minimum is observed again. When this effect is increased too much, the agent never finds a stable policy despite finding the optimum path many times.

In order to lower the training period, the sizes of the key and query matrices of the models used for Box-World have been reduced, and the optimizer learning rates have been increased. Apart from this, the gradient clipping value, which was not specified/used in the RN, was found to be highly effective in terms of the stability of the loss function. The loss function is quite unstable at high clipping values but converges very slowly at low values. The maximum of the learning rate value is selected as the range used in the RN. The reason for this is to achieve the highest speed training performance with tested parameters. In the experiments about the number of agents used in the A3C algorithm, the number of agents and the convergence of the policy were directly proportional as expected. This is because of the high variance of information that each agent collects in different environments [14]. The upper limit used in the number of agents is due to the hardware limitations.

Considering the attention weights of the RN, dense attention is generally seen between the agent, terminal, sword and pit objects. The reason for this is that all four objects are the objects that most affect the cumulative reward. When looking at Figure 7, it was seen that the reward values are in different scales for two configurations. This is because the configurations have a different number of distractor objects. Therefore, it is feasible to evaluate the environments within themselves. Although the standard deviation of RN is higher in both environments, the overall performance is still higher. However, compared to the Box-World environment, the performance difference between the two algorithms decreased considerably in the training episode interval. Although the RN algorithm performed relatively better in the first configuration, it appears that in the second configuration —higher in terms of complexity— their performance is quite close.

\begin{figure}
\centering 
\subfloat[Cumulative reward on RGW configuration 1.]{%
\includegraphics[clip,width=5in,height=2.02in]{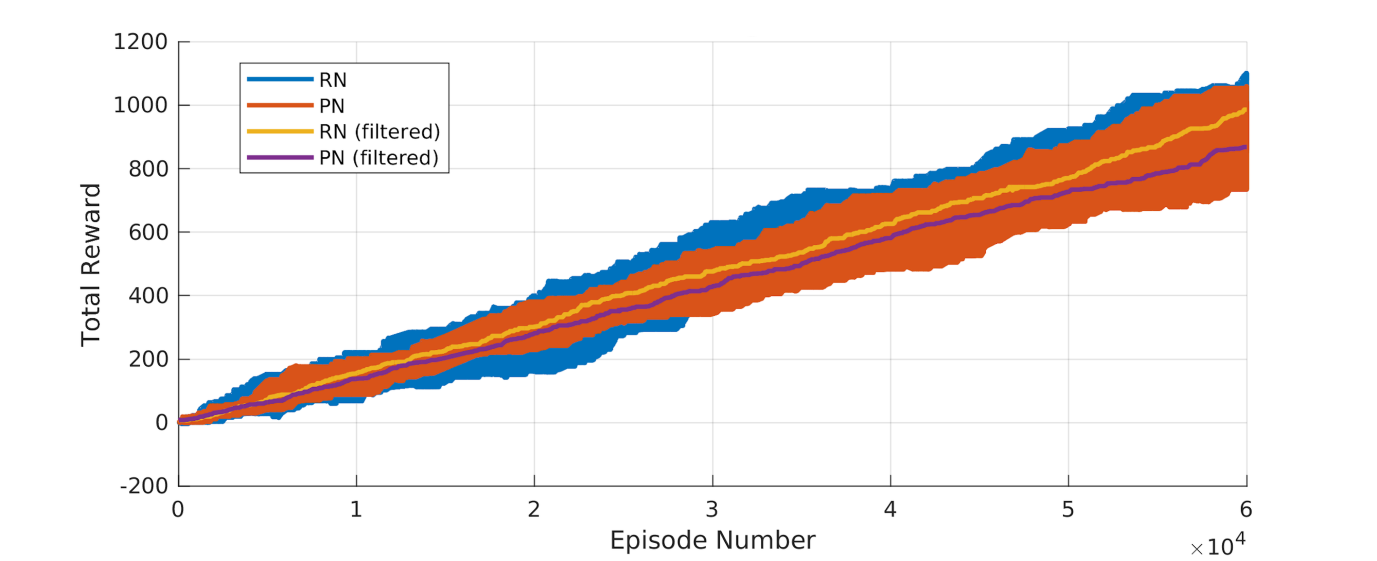}%
}

\subfloat[Cumulative reward on RGW configuration 2.]{%
\includegraphics[clip,width=5in,height=2.02in]{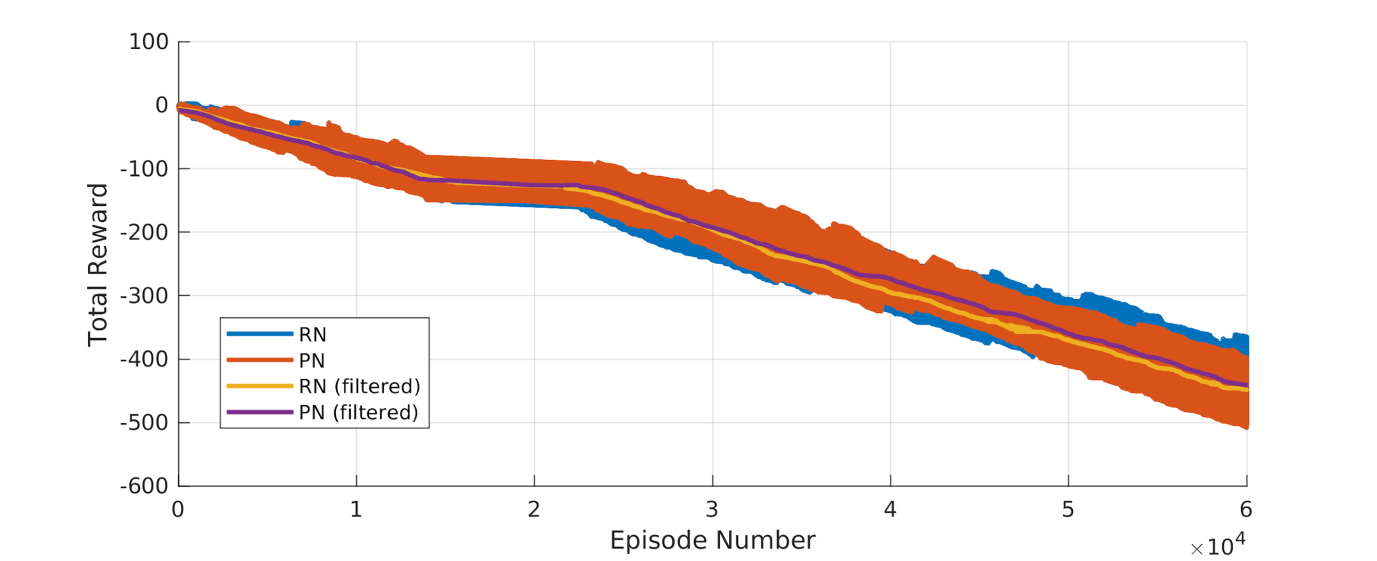}%
}

\caption{Relational-Grid-World (RGW) environment performances of PN and MHDPA (RN) modules.} 
\end{figure}

PrediNet Module was used for the first time for an RL problem and it is compared with RN as a baseline algorithm in the two different environments. As a result of this comparison, it was seen that the PrediNet algorithm performed closely with the RN algorithm in the episode range, where the agents were trained in the experiments. As expected, the RN module, which is a more complex network, converged to a global maximum in a longer wall-clock time than PrediNet. Therefore, it has been observed that PrediNet can be preferred in cases where there are limitations in computation. Apart from this, the relations between the objects in the environment are extracted explicitly with the PrediNet module. Extracted relational information is used by the agent for the production of policy logits and value estimation. Unlike RN, the PrediNet module, which produces explicit information, is also preferable in this respect. Apart from this, the RGW environment has been presented to measure the relational capacity of the RL agent and to create different decision-making problems. In order to reach the optimum solution path, if necessary, the related objects in the environment must be taken in the correct order. Distraction effect was created by using these objects close to other objects in the environment. Since the reward function used for the RGW was created by establishing an analogy with the baseline environment, it was sufficient to make small changes in the hyperparameter sets of the tested agent architectures. By testing the same architectures in both environments, the RGW environment was found to have sufficient measurement capacity.

\section{Conclusion}
Reinforcement learning agents that are designed using neural networks may not work in a semantically similar environment to the trained environment with different visual properties. This problem leads us to the generalizability and interpretability problems of the RL agents. There are several attempts to boost statistical Deep RL agent networks in order to avoid these problems with symbolic operations, such as using relational information between the environmental objects. These operations can give strong abilities that can be adapted to different RL problems. In this work, we introduced a novel reinforcement learning architecture that uses relational representations between environment objects in order to solve a sequential decision-making problem. In this model, we used the PrediNet architecture in an asynchronous actor-critic  (A3C) framework. Then, we compared the relational representation performances of PrediNet with the Multi-Head Dot Product Attention Network (MHDPA) module. In the results, we found that the PrediNet module network reaches close performance with MHDPA in a limited number of episodes. Unlike MHDPA, PrediNet can establish explicit numerical relational information for different relational representations between the objects. We used different Box-World (as a baseline) and our Relational-Grid-World (RGW) environment configurations for the experiments on two modules. We proposed the novel RGW environment, which contains eight objects with different functions in order to measure the RL agent's relational representation capabilities. In the experiments conducted in the RGW environment, it was seen that the relational modules could establish a direct connection between the objects with an expected difficulty. Therefore, RGW can be a useful tool in order to make the measurement of relational representation methods.

As a future work, we aim to visualize the explicit relational information created by PN as in the MHDPA module and increase the interpretability of the network. Increasing generalizability through relational information and symbolic operations is our ultimate aim. Also, we plan to test the modules in a higher variance state-space by generating the RGW environment procedurally.


\appendix

\begin{table}[h]
\renewcommand{\arraystretch}{1.3}
\caption{The hyperparameters of the agent architecture}
\label{table6}
\centering
\begin{tabular}{lc}
\hline
Parameter & Value\\
\hline
RL-method & A3C w/ 12 actors\\
Gamma & 0.99 \\
Entropy weight & c \\
Maximum episode length & L\\
\hline
Input shape & n x n x 1  \\
\hline
CNN1 Output Channels & 12 \\
CNN1 Kernel Size & 2 \\
CNN1 Activation & ReLU \\
CNN1 Stride	& 1 \\
CNN2 Output Channels & 24 \\
CNN2 Kernel Size & 2 \\
CNN2 Activation & ReLU \\
CNN2 Stride	& 1 \\
\hline
Module input size & n x n x 26 \\
\hline
MHA number of heads & 4 \\
MHA key / query size & g \\
MHA value size	& 64 \\
MHA pooling strides &  1 x 1 \\
MHA output size	& 26 \\
\hline
PN number of heads & 4 \\ 
PN key / query size & g \\
PN relations & 8 \\
PN comparator & Vector Difference \\
PN output size & 26 \\
\hline
FC1 Output &  256 w/ ReLU \\
FC2 Output & 128 w/ ReLU \\
FC3 Output & 64 w/ ReLU \\
\hline
FC4 Output (policy logits)	&  4 w/ Softmax \\
FC4 Output (value estimation) &  1 w/ None \\
\hline
Buffer size	& 40 \\
Optimiser & RMSprop \\
Learning rate  & e \\	
Optimiser momentum & 0 \\
Optimiser epsilon & 0.1 \\
Optimiser decay	& 0.99 \\
Gradient clip & 400\\
\hline
\end{tabular}
\end{table}

\begin{table}[htp]
\renewcommand{\arraystretch}{1.3}
\caption{Environmental variables}
\label{table6}
\centering
\begin{tabular}{cccccc}
\hline
Environment & L & n & g & c & e \\
\hline
Box-World  & 300 & 12 & 64 & 2e-4 & 0.01 \\
\hline
Relational-Grid-World  &  200 & 10 & 32 & 2e-3 & 0.02 \\
\hline
\end{tabular}
\end{table}

\end{document}